# Building with Drones: Accurate 3D Facade Reconstruction using MAVs

Shreyansh Daftry, Christof Hoppe, and Horst Bischof

*Abstract*— Automatic reconstruction of 3D models from images using multi-view Structure-from-Motion methods has been one of the most fruitful outcomes of computer vision. These advances combined with the growing popularity of Micro Aerial Vehicles as an autonomous imaging platform, have made 3D vision tools ubiquitous for large number of Architecture, Engineering and Construction applications among audiences, mostly unskilled in computer vision. However, to obtain high-resolution and accurate reconstructions from a large-scale object using SfM, there are many critical constraints on the quality of image data, which often become sources of inaccuracy as the current 3D reconstruction pipelines do not facilitate the users to determine the fidelity of input data during the image acquisition. In this paper, we present and advocate a closed-loop interactive approach that performs incremental reconstruction in real-time and gives users an online feedback about the quality parameters like Ground Sampling Distance (GSD), image redundancy, etc on a surface mesh. We also propose a novel multi-scale camera network design to prevent scene drift caused by incremental map building, and release the first multi-scale image sequence dataset as a benchmark. Further, we evaluate our system on real outdoor scenes, and show that our interactive pipeline combined with a multi-scale camera network approach provides compelling accuracy in multi-view reconstruction tasks when compared against the state-of-the-art methods.

## I. INTRODUCTION

Micro Aerial Vehicles a.k.a. Flying drones make themselves useful in a number of Architecture, Engineering and Construction (AEC) applications. With the price of the technology dropping fast, these firms are putting drones to work with increasing frequency. Building teams across the globe are now employing such drones as imaging platforms to simplify what is inherently a complex, messy process. Adding a simple passive camera technology makes these drones become invaluable: to fly through existing structures while returning rich, multi-layered data about the building information that can be used for creating 3D digital models of exterior facades, plan renovations via digital 3D simulation and even to automatically generate 3D printable pre-fabricated structures.

While the field of drones and robotics is yet emerging, image-based 3D reconstruction and modeling techniques have reached maturity. SfM is a well-studied topic and several methods have surfaced [2], [3] in the last few years, which demonstrate increased robustness, result in high quality and claim to have accuracy comparable to laser range

All authors are members or alumni of the Aerial Vision Group, Institute of Computer Graphics and Vision, Graz University of Technology, 8010 Graz, Austria. Shreyansh Daftry is also with Robotics Institute, Carnegie Mellon University, 15213 Pittsburgh, USA. This work was supported by Austrian Research Promotion Agency (FFG) in collaboration with Winterface GmBH. daftry@cmu.edu, {hoppe,bischof}@icg.tugraz.at

sensor systems at a fraction of the cost [4]. Though, these methods were designed to efficiently process thousands of unordered images from photo community collections and automatically construct 3D models [1], their implicit benefits have found popularity not only among researchers, but also lead to participation of an ever increasing population of end-users, who are not experts in computer vision, for applications like architectural reconstruction and scene documentation.

We agree that, given a set of images and method, a certain accuracy can be achieved that is comparable to lasers, as claimed by most recent work. However, it is also widely understood in the community that an arbitrary set of images can be found that would not meet that accuracy. Furthermore, for industrial applications like automatic facade reconstruction, the accuracy of the resultung 3D models are extremely crtitical and the current methods for image-based 3D reconstruction do not match required accuracy under unconstrained circumstances. The quality and completeness of 3D models obtained by SfM heavily depend on the image quality and image acquisition strategy. Since the current pipelines are designed mainly to handle large-scale unordered image datasets with very high redundancy, they often do not meet the demands of application specific scenarios where images are acquired deliberately for the reconstruction process. Hence, one needs to come up with novel measures for a predictable application of SfM in the most diverse circumstances. This work thus focuses on Structure-from-Motion for applications where images need to be deliberately acquired for the reconstruction process.

To recover an accurate and complete 3D model, the SfM process has several requirements on the input images: The viewing angle between two images may not be too large to allow feature matching, the view cones must overlap, the images have to be textured but the texture may not be repetitive and lighting hasn't changed too much between images. For a user it is impossible to estimate if the acquired images fulfill all demands. Another difficult question is the scene completeness, i.e. the coverage of the scene. Parts of the scene that are not captured with sufficient image overlap cannot be reconstructed. Since completeness depends on the required reconstruction resolution i.e. level-of-detail, and on the surface itself it is not possible to quantify the degree of completeness without prior information. In contrast, for a human it is relatively easy to answer this question by comparing a 3D model to the real world.

We propose to tackle this problem by integrating the acquisition process directly into the reconstruction pipeline. Instead of first acquiring a set of images and then processing the whole dataset in batch-based manner, we propose a closed-loop interactive processing that performs an incremental reconstruction in real-time. Our closed-loop approach first determines the camera pose of a new acquired image, updates the sparse point cloud and then extracts a surface mesh from the triangulated image features. This allows us to compute quality parameters like the Ground Sampling Distance (GSD) and the image overlap which are then visualized directly on the surface mesh.

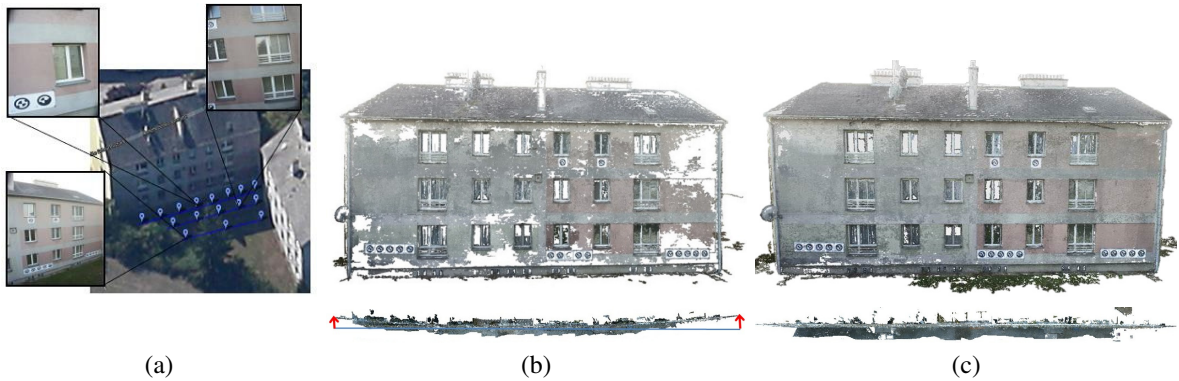

Fig. 1. (a) Graz500 Multi-scale Facade dataset. (b) Reconstruction of a facade shows that though visually correct in appearance, geometric inconsistencies (significant bends along the fringes) and holes in reconstruction are prevalent, even with state-of-the-art method [1]. (c) In comparison, using our multi-scale framework with a online feedback support results in a straight wall and complete scene reconstruction.

Furthermore, the intuitive difficulties of incremental SfM methods, such as drift, high resolution vs. accuracy tradeoff, etc. have not been solved completely (Fig. 1). The explanation for the cause of this drift lies in the actual camera network: as the scene is covered by fewer images towards the borders of the surveying area compared to the center of the object. Less image overlap leads to fewer image measurements per object point and thus causes the camera network to have fewer connections at the borders. This has the effect that the optimization of camera positions and 3D object points in the bundle adjustment is less constrained, thus the optimized positions can undergo larger changes.

To overcome this problem, we propose to adjust the image acquisition strategy to a multi-scale network and take images at different distances to obtain dense reconstructions while being more accurate. Our motivation for a multi-scale camera network also derives from our experience in 3D aerial mapping. Flying at different altitudes is a common approach in airborne photogrammetry to enhance the accuracy of I/O parameters. We use the same idea to ground level image-based modeling and propose to change the gold-standard from acquisition at a single depth to a multi-scale approach. It is the above considerations that motivate our contribution.

Our technical contributions are three-fold. First, we present a user-friendly method for accurate camera calibration using fiducial markers, an automatic geo-referencing technique and a fully incremental surface meshing method using a new sub-modular energy function, which are then integrated into a full real-time system. Second, we, propose a multi-scale camera network design to constrain the bundle block optimization to prevent scene drift caused by incremental map building, while obtaining high quality reconstruction. Third, we release the first multi-scale image sequence dataset along with complete ground truth to be available for the community as benchmark data.

## II. BENCHMARKING DATASET

We acquired a dataset (Graz500 Multi-scale Facade) with a multi-scale image sequence of an outdoor facade scene consisting of 500 images. For image acquisition we used a Falcon octocopter from AscTec as a flying platform. The MAV is equipped with a Sony NEX-5N digital camera. The APS-C CMOS sensor has an image resolution of 16.1 megapixels and is equipped with a fixed focus lens having a focal length of 16 *mm*. The overview of the multi-scale camera network design is depicted in Fig. 1. Images were acquired at different depths, heights and viewing angles to facade using the online feedback methodology described in the next section. The dataset thus also offers an opportunity for detailed analysis and evaluation of the various factors in image-based facade reconstruction. To the best of our knowledge, this is the only publicly available dataset with a structured multi-scale image sequence and is expected to serve as a benchmark data for the community.

We have acquired accurate terrestrial laser scanning (LIDAR) data, having a GSD of 1.5 *cm* and point measurement uncertainty of 2 *mm* using a Leica Total Station that will serve as geometrical ground truth to evaluate the quality of the image based reconstructions. The data was acquired in a single scan and hence does not involve any irregularities due to scan registrations. Note, that we do not treat the LIDAR data as an absolute ground truth reference. The LIDAR itself contains errors, which are estimated and translated to the image based acquisition. As a result, we evaluate image-based modeling only relative to its accuracy. Thus, to assess the achieved absolute accuracy in 3D, the facade (which is about 30m high and 50m long) is equipped with a reference network (17 fiducial targets) of Ground Control Points. The ground truth data for each GCP is measured using a theodolite and has an uncertainty of less than 1 *mm*.

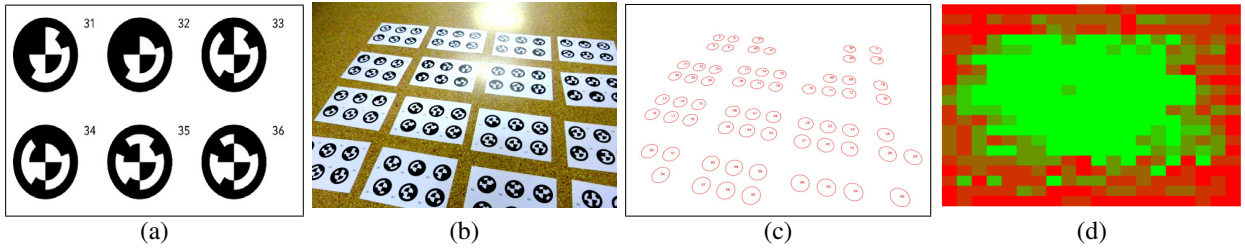

Fig. 2. Camera Calibration using fiducial markers. (a) Each marker encodes a unique ID. (b) A typical calibration image arbitrarily arranged in a 4x4 grid on the floor. (c) Reliably extracted markers with positions and ID. (d) Online feedback for feature point coverage.

## III. APPROACH

Our 3D facade reconstrcution pipeline follows the traditional approach and consists of four modules: image acquisition and camera calibration, multi-view sparse reconstruction, geo-registration by transforming the 3D model into a geographic reference coordinate frame and finally, a densification and surface meshing step. In order to realize the closed-loop schema that can give instant feedback about the contribution of a new image, all components have to work (a) in a fully incremental manner and (b) in real-time. Therefore, we present and employ novel interactive calibration and incremental sparse reconstruction schemes, an automated geo-referencing technique and a fully incremental surface meshing method, which are then integrated into a full real-time system.

### A. Accurate Camera Calibration

Accurate intrinsic camera calibration is critical to most computer vision methods that involve image based measurements. Traditional SfM pipelines such as Bundler, Agisoft, etc. employ a direct use of uncalibrated views for 3D reconstruction, and can inherently deal with a dataset having images taken at varying focal length, scale and resolution. However, in our experience, we have found that accuracy of Structure-from-Motion computation is expected to be higher with an accurately calibrated setup [5], [6]. In most of the calibration literature [7], [8], [9], a strict requirement on the target geometry and a constraint to acquire the entire calibration pattern has been enforced. This is often a source of inaccuracy when calibration is performed by a typical end-user. Additionally, these methods tend to fail when images are taken at considerably different distances to the object. Hence, aiming at the accuracy of target calibration techniques while factoring out image space variations due to occlusion, reflection, etc., we advocate the use of a recently proposed fiducial marker based camera calibration method [10].

The calibration routine follows the basic principles of planar target based calibration and thus requires simple printed markers to be imaged in several views (Fig. 2). Each marker includes a unique identification number as a machine-readable black and white circular binary code, arranged rotationally invariant around the marker center. A novel technique for robustly estimating the focal length is employed, where an error function is exhaustively evaluated to obtain a globally optimal value of focal length $f$ determining the calibration matrix $K$. In our findings, this methods works very robustly and performs much better for a multi-scale image sequence acquired at varying depths to the object, as compared to traditional methods that employ a non-linear minimization technique for intrinsic parameters estimation. In addition, staying with the thrust of this paper to give users a online feedback of the acquisition fidelity, we facilitated the method with an easy to use GUI with user-feedback about the feature point coverage (Fig. 2). There are significant qualitative and quantitative benefits of the presented calibration method towards a multi-scale robust image sequence.

### B. Real-time and Interactive SfM

Classically, batch-based SfM approaches assume spatially unordered images as input and therefore require several minutes or hours to determine the spatial ordering by constructing an epipolar graph [1]. Since the construction of the epipolar graph comprises the calculation of relative orientations between all image pairs, this is the most time-consuming task in batch SfM pipelines [11]. In our addressed problem, we can assume that a user does not acquire images in a totally random order. If we assume that a new input image $I$ has an overlap to an already reconstructed scene part, we can skip the epipolar graph construction and the SfM problem can be split into two tasks that are easier to solve: localization and structure expansion. More formally, given a freshly acquired input image $I$ and a reconstructed scene $M$, we find the position of $I$ within $M$ and finally, we expand the map $M$. The presented method is similar to visual SLAM, but it matches wide-baseline features instead of tracking interest points.

To calculate the pose of an image with respect to an SfM point cloud we follow the approach of Irschara *et al.* [12]. Given a new image $I$, we compare its visual appearance against all already reconstructed images using an efficient vocabulary tree approach [13] and results in a similarity score for each image. Then we match $I$ pairwise against the image features of the top n images with highest similarity score to determine feature correspondences. Since some of the features are already used for the triangulation of 3D points, we can establish 2D-3D image correspondences between $I$ and $M$. Given a set of 2D-3D correspondences and a calibrated camera, we solve the absolute pose problem robustly in a RANSAC loop [14]. If a valid position for $I$ is determined, this pose is refined by minimizing the reprojection error using non-linear optimization techniques.

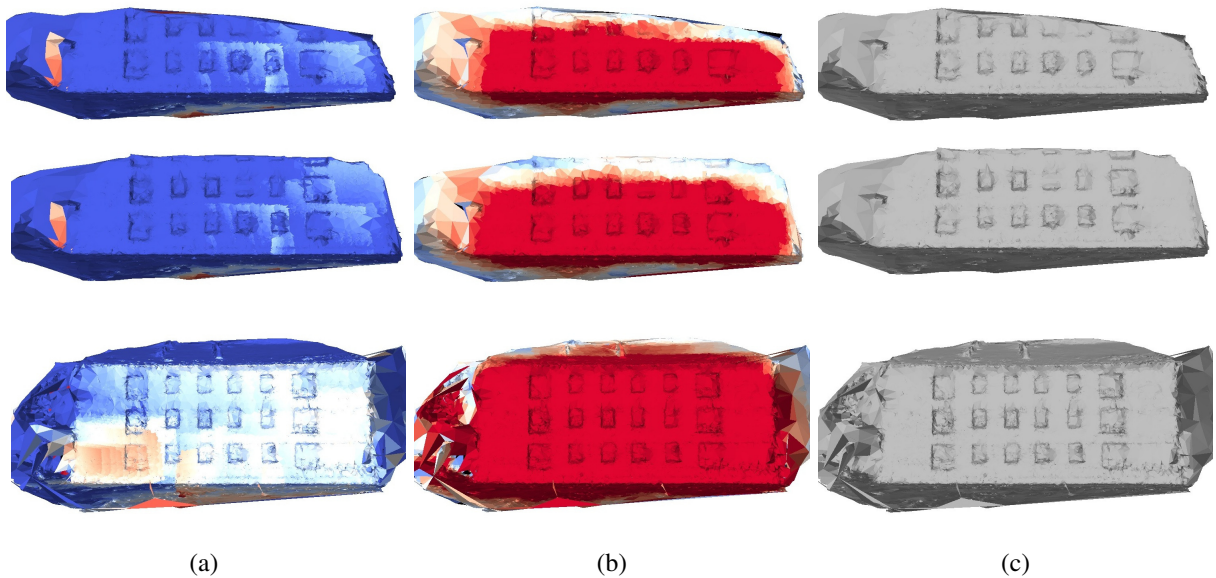

(a)          (b)          (c)

Fig. 3. Online feedback of the evolution of the reconstructed facade over time. The first column shows the color coding of the GSD where blue indicates a low resolution and red a high resolution. The images in the second column are the image overlap of the reconstruction. Red colored parts are seen by 30 and more cameras. The last column shows the reconstructed mesh without any color. The meshes are extracted after 50, 100 and 200 integrated images. The visualization supports the user to recognize which parts of the object have been reconstructed reliably.

If we cannot localize $I$ within $M$, this is reported to the user instantly. Hence, they can directly know that $I$ could not be aligned within the map and are asked to take a new picture. Since the orientation of $I$ is known and we already have image correspondences available from the previous step, we can easily triangulate new 3D points. To reduce the number of outlier matches, we perform epipolar filtering before triangulation. For bootstrapping the initial map $M$, we require two images taken from different viewpoints, and perform brute-force feature matching.

### C. Incremental Surface Reconstruction

The basic principle of most existing methods for extracting surfaces from sparse SfM point clouds is based on the Delaunay triangulation (DT) of 3D points, and then to robustly label the tetrahedra into free and occupied space using a random field formulation of the visibility information. The surface is then extracted as the interface between free- and occupied space. Such traditional methods for robust surface extraction [15] are not well suited for incremental manner as their computational costs scale exponentially with the increase in the number of 3D points. In contrast, existing solutions for incremental surface reconstruction [16], [17] are either based on a strong camera motion assumption or they are prone to outliers [18].

We present a new method to incrementally extract a surface from a consecutively growing SfM point cloud in real-time. Though the core idea of our method remains the same, we propose a new energy function that allows us to extract the surface in an incremental manner, i. e. whenever the point cloud is updated, we adapt our energy function. Given a set of tetrahedra $V$ obtained by the DT of the point cloud, we define a random field where the random variables are the tetrahedra of $V$. Our goal is to identify the binary labels $\mathscr{L}$ that give the maximum a posteriori (MAP) solution for our random field, analyzing the provided visibility information $\mathscr{R}$. The binary labels specify if a certain tetrahedron $V_i \in V$ is free- or occupied space. To identify the optimal labels $\mathscr{L}$, we define a standard pairwise energy function:

$$E(\mathscr{L}) = \sum_i (E_u(V_i, \mathscr{R}_i) + \sum_{j \in \mathscr{N}_i} E_b(V_i, V_j, \mathscr{R}_i)) \quad (1)$$

where $\mathscr{N}_i$ denotes the four neighboring tetrahedra of the tetrahedron $V_i$ and $\mathscr{R}_i$ is a subset of $\mathscr{R}$, consisting of all rays connected to the vertices that span $V_i$.

For defining the unary costs $E_u(V_i, \mathscr{R}_i)$ we follow the idea of the Truncated Sign Distance function (TSDF) that the probability that a certain tetrahedron $V_i$ is free space is high, if many rays of $\mathscr{R}_i$ pass through $V_i$. Therefore, we set costs for labeling $V_i$ as occupied space to $n_f \alpha_{free}$, where $n_f$ is the the number of rays of $\mathscr{R}_i$ that pass through $V_i$. In contrast if $V_i$ is located in extend of many rays of $\mathscr{R}_i$ the probability is high that $V_i$ is occupied space. For this reason, the costs for labeling $V_i$ as free space are set to $n_o \alpha_{occ}$, where $n_o$ is the number of rays in front of $V_i$. Fig. 4 illustrates the unary costs for a small example. Here, $n_f$ is 1 since only the light blue ray passes $V_i$ and $n_o$ is 3 because $V_i$ is in extend of the three green rays. The red rays do not contribute to the unary costs.

For the pairwise terms we assume that it is very unlikely that neighboring tetrahedra obtain different labels, except for pairs $(V_i, V_j)$ that have a ray through the triangle connecting both. Let $R_k$ be a ray of $\mathscr{R}_i$ that passes $V_i$. If $R_k$ intersects the triangle $(V_i, V_j)$, $E_b(V_i, V_j, \mathscr{R}_i)$ is set to $\beta_{vis}$. Triangles $(V_i, V_j)$ that are not intersected by any ray of $\mathscr{R}_i$ are set to $\beta_{init}$. Fig. 4 shows the pairwise costs in an example. Fig. 4 visualizes the graphical model of the energy function for a small example.

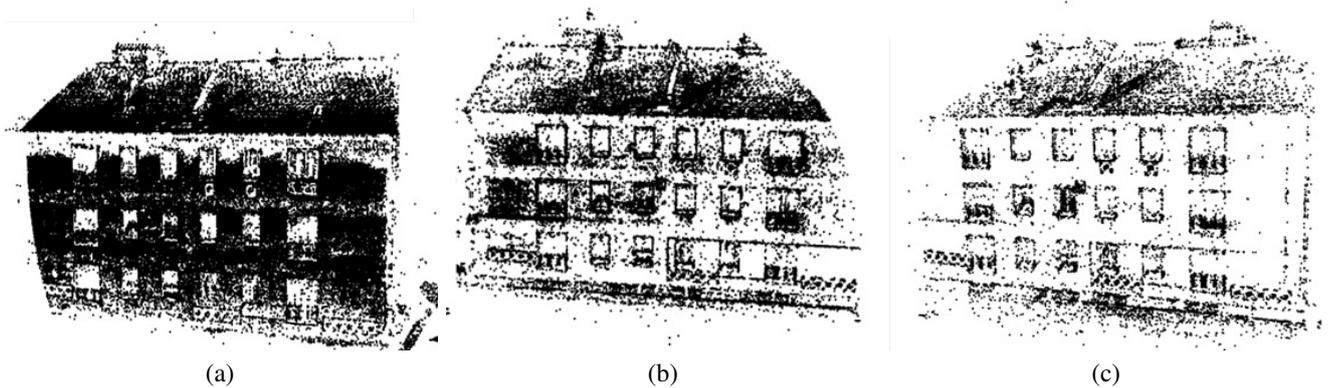

(a)             (b)             (c)

Fig. 5. Scene completeness for SfM results from individual row subsets at (a) Distance: 4m (b) Distance: 6m and (c) Distance: 10m.

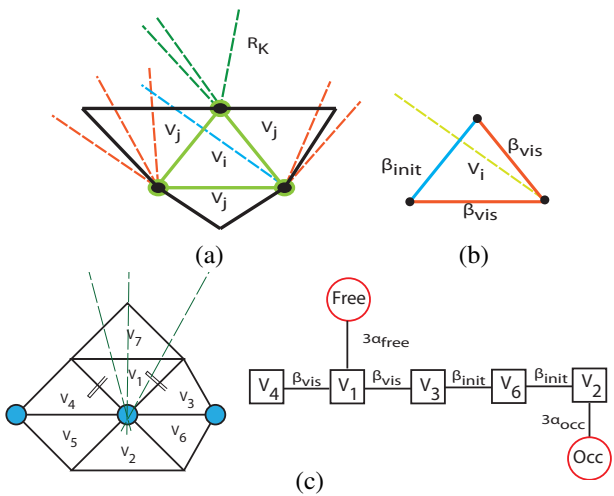

Fig. 4. (a) For defining the unary term for a specific tetrahedron $V_i$ we only analyze rays (dashed lines) connected to vertices that span $V_i$. (b) For the pairwise term we only consider rays that pass through the tetrahedron and that are connected to the tetrahedron vertices. (c) Graph representation of the energy function. The pairwise weights that are not shown are set to $b_{init}$.

Having defined all terms for our random field formulation, we are then able to derive a globally optimal labeling solution for our surface extraction problem using graph cuts. This is possible as our priors are submodular. However, to enable the method to work in an incremental fashion, we employ the dynamic graph cut of Kohli *et al.* [19] instead of minimizing the updated energy with a standard graph cut. This allows an efficient minimization of a series of similar random fields by re-using the previous solution.

To guide the user throughout the acquisition, we visualize the current, Ground Sampling Distance (GSD) and image redundancy as quality indicators [18] on the surface model as shown in Fig.3. Quantitatvely, our method achieves the same accuracy as state-of-the-art methods but reduces the computational effort significantly. The difference in computational effort is mainly caused by the definition of the energy function. While [15] has to perform a full raycast for each ray, we only have to identify the tetrahedra in front and behind the vertex and the first triangle that is intersected by the ray, and while their optimization requires 740 ms, our energy is fully optimized in 430 ms. Hence, the combination of the dynamic graph cut with our new energy formulation allows us to extract the surface from an increasingly growing point cloud nearly independent of the overall scene size in almost real-time.

### D. Multi-scale Camera Network

In 3D reconstruction literature, the distance to the reconstruction object has always been considered an important and contributing factor but seldom has been studied in an empirical way. It is widely accepted in the robotics and vision community that the closer we go to the image greater information is gained and accuracy is improved. However, our experience with structure from motion has shown that the contrary is true. We performed a systematic study on the ground control point accuracy with respect to distance of image acquisition from facade and ground sampling distance. Our facade dataset was further quantified into 3 row-subsets based on the distance of acquisition from close to distant (4 *m*, 6 *m* and 10 *m*), and reconstruction was performed on each subset independently using the proposed pipeline.

*1) Scene Completeness Vs. High Accuracy:* It can be observed from the results in Table I, the mean absolute error decreases significantly as we go further from the facade. This is contrary to the belief that the closer one gets to the object i.e. the higher the resolution the greater will be the accuracy. Thus after an exhaustive evaluation and study of various parameters we can state that the influence of the geometric configuration of the multi-view camera network on the resulting accuracy is higher than the influence of redundancy in image acquisition and there is a significant accuracy gain as we go away from the facade. This is possibly due to the strong drift affect caused in the camera pose estimation when the distance between the camera and the object is very small. However, we also observe that as we go closer to the facade the scene completeness is greatly improved as can be seen in Fig. 5. This is because a greater

| | Mean Error (mm) | | |
|---|---|---|---|
| Distance | Ours | Bundler | AgiSoft |
| Near Row (4m) | 45.2 | 51.3 | 57.1 |
| Middle Row (6m) | 23.1 | 27.2 | 32.3 |
| Far Row (10m) | 5.7 | 11.2 | 16.1 |

TABLE I

ACCURACY RESULTS ON INDIVIDUAL ROW SUBSETS.

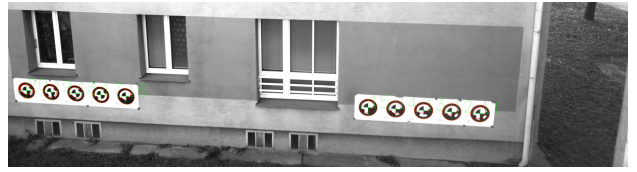

Fig. 6. Automatically detected ground control points with plotted marker centers and corresponding marker IDs.

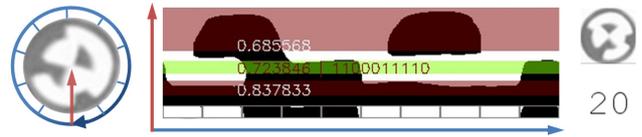

Fig. 7. Histogram for an unrolled circular marker and rotation invariant binning of the code stripe. The numbers from top to bottom indicate the probabilities for center, binary code and outer circle. The marker with ID 20 has been successfully decoded.

number of finely textured featured points are only visible in the closeby images due to a higher GSD. It can be thus concluded that there is a trade-off between accuracy and completeness as we change the distance between the image acquisition and facade. We can generalize this as a systematic behavior as they can also be consistently observed in the standard software packages. Hence, we infer that a model incorporating the knowledge of this trade-off could help in improving the metric accuracy of the final reconstruction.

*2) Constrained Bundle Block Optimization:* A way to avoid systematic errors arising from the deformation of the image block is to introduce the prior knowledge of this behavior in bundle adjustment. The global optimization of camera positions and 3D object point positions in the Bundle Adjustment (BA) is carried out based on Googles Ceres Solver for non-linear least squares problems [20]. For this purpose, we modify the cost function from the original bundling step to include another cost term based on distance of the image from the scene of interest besides the mass of object points from natural landmarks from SIFT. In particular, we penalize the reprojection error of the feature points in image space with a Huber error norm [21]. Furthermore, we let the initial camera parameters for intrinsics and lens distortion be commonly refined and corrected for all cameras in the bundle adjustment step. Thus multi-scale integration distributes the residual errors equally over all markers and allows 3D reconstructions with very low geometric distortions, even for elongated objects of large extent. This can be observed from the experimental results shown in Sec. IV

*E. Automatic Geo-Referencing*

The reconstruction and external orientation of the cameras recovered using the described pipeline so far is initially in a local Euclidean coordinate system and only up to scale and therefore not metric. However, for applications in facade reconstruction, the absolute position accuracy of the measured object points is of interest [22]. In addition, we want the created 3D model correctly stored and displayed in position and orientation in its specific geographic context. Based on known point correspondences between reconstructed object points and ground control points (GCPs), we first transform the 3D model from its local source coordinate frame into a desired metric, geographical target reference frame using similarity transform.

*1) Marker-based Rigid Model Geo-Registration:* To facilitate automation and to avoid erroneous point correspondences by manual control point selection, the association of point correspondences between image measurements and ground control points is encountered again using fiducial markers introduced for camera calibration. A requirement for full automation is that markers are detected robustly and stable in at least two images of the dataset and are clearly identified individually by their encoded ID. The detection also needs to work reliably from different flying altitudes and distances from the object.

Instead of paper print outs, we make use of larger versions ($\sim$50 cm diameter) of the markers printed on durable weather proof plastic foil to signal reference points in the scene used as GCPs. The markers are flexible, rolled up easy to carry, though robust and universally applicable even in harsh environments.

The markers are equally distributed in the target region and placed almost flat on the ground or attached to a facade. The 3D positions of the marker centers are then measured by theodolite, total station or differential GPS with improved location accuracy (DGPS), which is the only manual step in our reconstruction workflow besides image acquisition. All input images are then processed for marker detection. After thresholding and edge detection, we extract contours from the input images and detect potential markers by ellipse fitting. The potential markers are then rectified to a canonical patch and verified by finding circles using Hough transform. If the verification is positive, we sample the detected ellipse from the original gray scale image to unroll it and build a histogram (Figure 7). In the thresholded and binned histogram we extract the binary code of the marker if the code probability is high. The marker ID is obtained by checking the code in a precomputed lookup table.

The detected ellipse center describes the position of the image measurement of the respective marker (see Figure 6). By triangulating multiple image measurements of one and the same marker seen in several images, we calculate its 3D object point position in the local reference frame of the model. The markers can be directly matched and automatically associated with their corresponding ground control reference points as long as they share the same ID. Once, corresponding point pairs have been established between model and reference, they are used to calculate the transformation parameters. This results in a geo-registered model of the object.

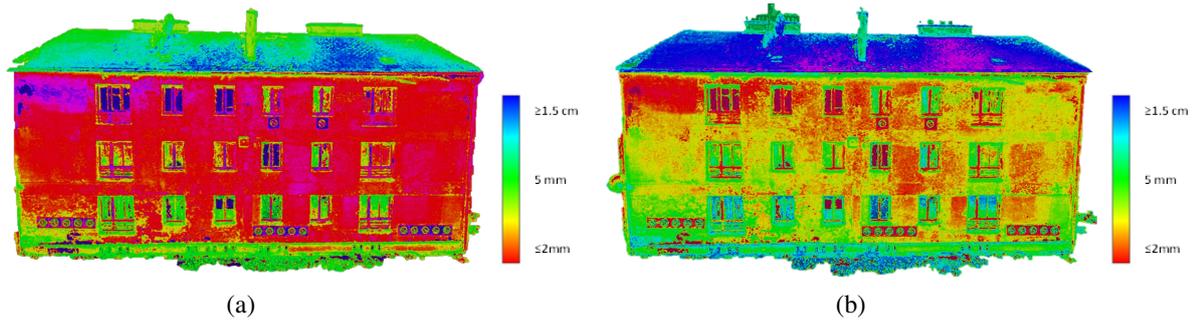

Fig. 8. Color coded dense 3D point clouds based on Hausdorff distance obtained using (a) Ours and (b) Bundler. Note: Best seen in color.

## IV. Performance Evaluation and Results

In this section we analyze the performance of our proposed approach for acuurate 3D facade reconstruction based on a multi-scale image sequence obtained from a a micro-aerial vehicle. All experiments were conducted in an uncontrolled environment to simulate conditions similar to that experienced by a common user.

### A. Accuracy Comparison to Ground Truth.

In order to give a full quantitative evaluation of the influence of our interactive SfM framework on reconstruction accuracy we compare our methodology to state-of-the-art pipelines using ground truth 3D data. The Bundler (open-source) [1] and Agisoft (commercial) software packages were used as our primary reference, as they represent the most popular methods for SfM within the computer vision community.

*1) Absolute Error.:* In literature, the reprojection error of an object or feature points has often been used as an adequate measure for evaluating the accuracy of the SfM reconstruction. However, for most end-user applications the accuracy of the 3D reconstructed object points is of prime interest. We thus perform a point-wise comparison and evaluation. First, we calculate the absolute mean over all the points as the 3D eurclidean distance between the corresponding ground control point and the reconstructed and geo-referenced point. The results of the experiment have been shown in Table II. It has to be noted that the results of other pipelines have been calculated on the dataset acquired without our online feedback framework.

*2) Relative Error.:* Next, we calculate the one way Hausdorff distance (similarity measure) between the reconstructed point cloud (densification using PMVS [23]) and the point cloud obtained from the Laser Scanner. The number of points in the reconstructed point cloud was about 9 million, with a GSD of 1 *mm*. Similar steps were performed for the sparse point cloud obtained from the Bundler software. The reconstructed point clouds were then color coded based on the Hausdorff distance and the results have been shown in Fig. 8.

|  | Mean Error (mm) | | |
|---|---|---|---|
| **Image Acquisition Method** | **Ours** | **Bundler** | **AgiSoft** |
| Standard BA | 23.1 | 27.2 | 32.3 |
| Multi-scale BA | **9.1** | 15.5 | 21.6 |
|  | Time Taken (sec) | | |
| **Time Performance** | **Ours** | **Bundler** | **AgiSoft** |
| SfM | **880** | 6220 | 6455 |

TABLE II
(A) Accuracy comparison w.r.t ground truth data. (B) Time Performance

It can be clearly seen that using our method the absolute mean error for reconstructed GCPs is in the range of 9 *mm* and overall relative accuracy of 90 % of the facade is within 2 mm error range with respect to the laser point cloud, which is within the uncertainty range of the total station. A closer inspection reveals that the high errors are only along the sections of the point cloud missing in the laser scanner such as roof, missing window panes, etc. Thus, we observe that our method considerably outperforms the state-of-the-art methods in both the absolute and relative error analysis to get highly accurate results comparable to the uncertainty of the laser point cloud, even when performed by an inexperienced end-user.

### B. Time Performance compared to State-of-the-art

To evaluate the performance of online SfM approach, we compare running time of presented online SfM to a state-of-the-art batch-based SfM approach. For both methods, we use 5000 SIFT features per image that have largest scale. The features are extracted by the SIFTGPU implementation. Our approach requires 880 seconds to process the 500 images from the dataset, which is 7.1 times faster than Bundler. On average, our approach requires 1.75 seconds to integrate a new image into the structure and to extend the map. This is within the latency of the time constraints of the MAV to transmit back a new image from the next possible camera network position, and hence we can conclude that the online SfM method is approximately in real-time.

*C. Multi-scale Camera Network Benefits*

We extend our evaluation to quantitatvely and qualitatively assess the benefits of multi-scale camera network based acquisition when applied to incremental 3D reconstruction methods. Experiments on accuracy evaluation similar to last section is performed, with and without a multi scale network based bundle adjustment. The results can be seen in Table II. We observe that the proposed multi-scale camera network framework using the constrained bundle block formulation helps to overcome drift. It facilitates accurate reconstructions without compromising on scene completeness. The qualitive benefits on the geometric fieldity of reconstruction has been shown in Fig. 1. As a ground truth, we know that the reconstructed wall of the facade should be straight. However on a detailed inspection, we can clearly see that the reconstructed wall suffers from significant bending using a uni-scale acquisition approach, owing mainly to the drift due to map building in a incremental SfM framework. In contrast, the use of a multi-scale approach helps constrain the bundle block from deformation due to error accumulation and consequently results in accurate and complete reconstruction.

V. CONCLUSION AND FUTURE WORK

In this paper we have revisited and improved the classic incremental 3D reconstruction framework. Our technique enables real-time computation of structure from motion and provides users with online feedback to ensure that all relevant parts of the environment have been captured with enough overlap and with the desired resolution. Further, we also present a novel framework to modify the image acquisition to acquire images at various depths. This formulation helps constrain the error accumulation due to drift, which is inherent in incremental mapping methods. The overall framework greatly improves the reliability of image-based 3D reconstruction when image acquisition is done deliberately. We show that combining these technologies with developments in drone technology can return micrometrically accurate data that has immense applications in Architecture, Engineering and Construction domains.

In the future, we will extend our method by adding a path planner to the approach, making it suitable for autonomous image acquisition and active multi-scale camera network design. Even without this feature, our method supports users to judge the quality and the completeness during image acquisition on site and makes the eventual reconstruction result predictable. Furthermore, the software for the proposed interactive calibration approach, the benchmarking dataset for multi-scale image sequence along with ground truth (Laser Scans and GCP positions in global coordinates) will be made publicly available, so that a wide audience can benefit from our findings.